\documentclass[10pt,twocolumn,letterpaper]{article}

\usepackage[pagenumbers]{cvpr} 

\usepackage{graphicx}
\usepackage{amsmath}
\usepackage{amssymb}
\usepackage{booktabs}

\usepackage[accsupp]{axessibility}  

\usepackage[pagebackref,breaklinks,colorlinks]{hyperref}

\usepackage[capitalize]{cleveref}
\crefname{section}{Sec.}{Secs.}
\Crefname{section}{Section}{Sections}
\Crefname{table}{Table}{Tables}
\crefname{table}{Tab.}{Tabs.}

\def\cvprPaperID{4} 
\def\confName{CVPR ABAW}
\def\confYear{2022}

\begin{document}

\title{The Best of Both Worlds: Combining Model-based and Nonparametric Approaches for 3D Human Body Estimation}

\author{Zhe Wang\\
University of California, Irvine\\
{\tt\small zwang15@ics.uci.edu}
\and
Jimei Yang\\
Adobe Research\\
{\tt\small jimyang@adobe.com}
\and
Charless Fowlkes\\
University of California, Irvine\\
{\tt\small fowlkes@ics.uci.edu}
}
\maketitle

\begin{abstract}
  Nonparametric based methods have recently shown promising results in reconstructing human bodies from monocular images while model-based methods can  help correct these estimates and improve prediction. However, estimating model parameters from global image features may lead to noticeable misalignment between the estimated meshes and image evidence. To address this issue and leverage the best of both worlds, we propose a framework of three consecutive modules. A dense map prediction module explicitly establishes the dense UV correspondence between the image evidence and each part of the body model. The inverse kinematics module refines the key point prediction and generates a posed template mesh. Finally, a UV inpainting module relies on the corresponding feature, prediction and the posed template, and completes the predictions of occluded body shape. Our framework leverages the best of non-parametric and model-based methods and is also robust to partial occlusion. Experiments demonstrate that our framework outperforms existing 3D human estimation methods on multiple public benchmarks. 
  
\end{abstract}


\begin{figure}
\centering
\includegraphics[width=0.99\linewidth]{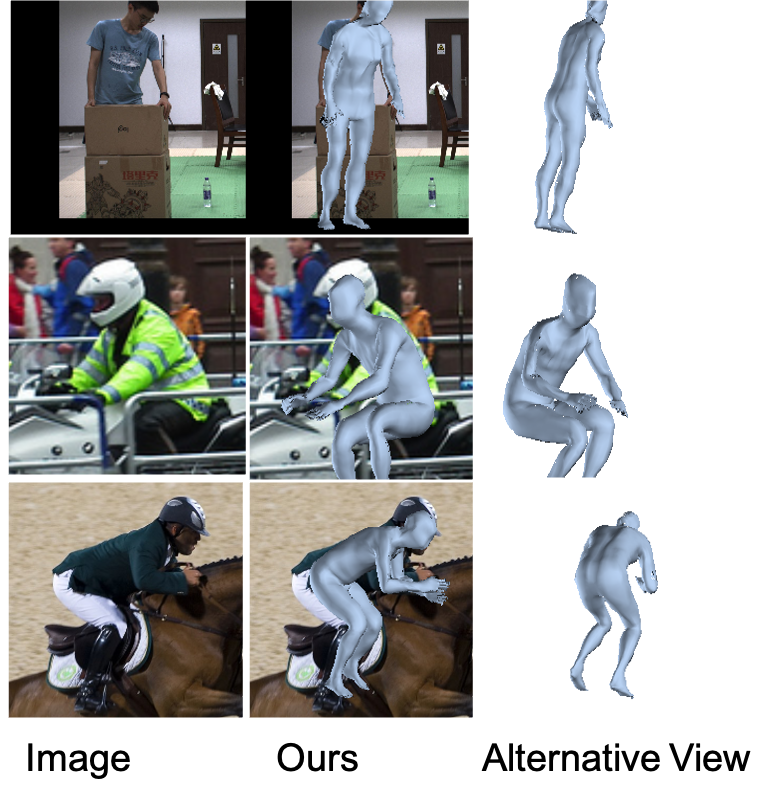}
\caption{{From left to right: Original image our mesh prediction overlay and alternative views mesh visualization. Images are from 3DOH \cite{3DOH},  and LSP \cite{lsp}  datasets. (Best viewed in Color)}}
\label{fig:firstimage}
\end{figure}

\section{Introduction}


The 3d estimation of the human body pose and shape from a monocular image is a fundamental task for various applications such as VR/AR, virtual try-on, metaverse and animations. It is challenging mostly due to the depth ambiguity and lack of evidence from single image.  There are several ways to solve this ambiguity such as leveraging multi-view or video data to fuse image evidence from more images and infer occluded parts. For the case of single images, researchers used parametric models such as SMPL \cite{smpl} to fit 2D image evidence \cite{spin} or use human pose prior \cite{HMR,SMPL-X:2019,kocabas2019vibe} to penalize problematic human pose / mesh prediction in combination with modern deep learning techniques. However, these model-based methods are prone to produce corrupted results when severe occlusion happens. 

Nonparametric methods use non-compressed representations like voxels \cite{volumetric}, heatmaps \cite{I2L} and joint location \cite{integral,Meshtransformer,meshgraphormer} as the target for modern deep learning. However, to estimate dense meshes they are computationaly expensive and consume lots of memory. They either use integral methods to estimate normalized joint location \cite{I2L} or simplify meshes \cite{meshgraphormer} to reduce the number of vertices. Without post-processing, these methods also generate qualitatively non-pleasing results. The dense correspondence methods \cite{cvprooh,decomr,pamigcndensepose}, which are based on template SMPL human mesh surface and have been proven for various tasks. 

Connecting nonparametric methods and model-based methods is hard due to the difficulty in localizing the corresponding feature. \cite{holopose,spatialarrangements,I2L} utilize bounding boxes or keypoints location to find the related features to estimate necessary SMPL parameters. While \cite{pare,Meshtransformer} learn the feature-parameter correspondence (attention) implicitly through neural networks. \cite{decomr,pymaf} consider the correspondence between the mesh representation and pixel representation based on human surface mapping (UV coordinate system). However, they estimate the SMPL parameter through a light weight FC network and treat this simple optimization process as a post process. Their methods also do not convey the advantages of nonparametric methods such as robustness to occlusion.

To leverage the advantages from both worlds, we propose a 3d human body estimation framework that consists of three modules: Dense Map Prediction module (\textit{DMP}), Inverse Kinematics module (\textit{IK}) and UV Inpainting module (\textit{UVI}). \textit{DMP} explicitly predicts per-pixel human 3d joint location, 3d surface location in root relative coordinates, 3d displacement between the joint location and surface location, and also predicts UV coordinates which represent the human surface in a 2D grid. This module is robust to partial occlusion when predicting joint, as all the image evidence belongs to this part will contribute to the prediction explicitly. \textit{IK} module connects the nonparametric prediction to model-based method.  We first warp the DMP dense prediction to UV space and get the joint prediction based on the part-segmentation in UV space. Then we use a two-stage multi-layer perceptron, where the first stage inpaints and refines the joint prediction, while the second stage estimates SMPL parameters and eventually produces a posed mesh. With all the predictions in UV space from \textit{DMP} and \textit{IK}, \textit{UVI} inpaints and refines the 3d body pose and mesh in UV space.

\noindent In summary, our contributions are three fold:
\\
$\bullet$ We propose a 3d body estimation framework from single image that seamlessly leverages the best of the both worlds (model-based and nonparametric). 
\\
$\bullet$ The method is robust to occlusions and can self-correct wrong poses from Dense Map Prediction module.
\\
$\bullet$ We achieve state-of-the-art performance on H36M and 3DOH datasets. 

\section{Related Work}

\paragraph{3D human shape estimation from monocular images}  SMPL \cite{smpl} has been widely used for 3D human mesh reconstruction. To boost its power in practice, a number of deep learning frameworks have been proposed by using SMPL as regression targets \cite{HMR,spin, SMPL-X:2019,decomr,I2L,pchmr}. \cite{HMR} regresses SMPL parameters directly from input images by end-to-end training. Following this research direction, \cite{I2L} add spherical Gaussian attention joint based on initial joint estimation, and the use the the attended feature to learn the vertices location. \cite{spin} combine learning and optimization\cite{SMPL-X:2019} in the same framework but cannot handle occlusions. \cite{decomr} uses the template UV mapping from SMPL and transforms 3d mesh reconstruction to decomposed UV estimation and position map inpainting problems. However, the way to get 3d human joint from SMPL mesh is based on the pre-trained joint regressor, which will induce intrinsic errors and usually does not generalize to other datasets.

\paragraph{3D human pose estimation from monocular images} Deep learning approaches
have shown success in regressing 3D pose from
a single image \cite{rootnet,LCRnet++,xiao2018simple,simple,semanticsgcn,volumetric,gpa,haoyubmvc2021,cross,dynamicpose}. Basically, most current models can be categorized into two frameworks. The first is to directly estimate 3D pose from images, based on volumetric representation \cite{volumetric,rootnet}. But these approaches may involve in high memory consumption and complex post-processing steps. Based on the explosive improvement
in 2D pose estimation \cite{xiao2018simple}, another framework is to estimate 2D pose from images and then lift 2D pose to 3D pose \cite{semanticsgcn,simple}. Since these approaches take 2D joint locations as input, 3D human pose estimation simply focuses on learning depth of each joint. This releases learning difficulty and
leads to better 3D pose. However, there are few methods on systematically handling occlusion in the first framework while the second framework cannot recover information if the joint detector fails. Additionally, how to get human surfaces from the joint prediction remains a problem. 

\paragraph{Inverse Kinematics} The inverse kinematics (IK) problem has been extensively studied in robotics \cite{ik1,ik3} and graphics \cite{ik2} and its techniques have been used in 3d human pose estimation \cite{motionretarget,HybrIK,pare,zhou2021monocular,zhou2020monocular}.  Numerical solutions \cite{ik1,ik2,ik3} rely on time-consuming iterative optimization.  \cite{motionretarget} uses temporal sequence to resolve IK ambiguity. \cite{HybrIK} decomposes the IK rotation to the product of swing rotation and twist rotation and solve swing rotation analytically from predicted joint locations. Feed forward solution like \cite{zhou2020monocular,zhou2021monocular} propose BodyIKNet to regress SMPL \cite{smpl} pose and shape parameters from 3d joint location, 
However, it leads to a sub-optimal solution when partial occlusion happens.

\begin{figure*}[t]
\begin{center}
   \includegraphics[width=1\linewidth]{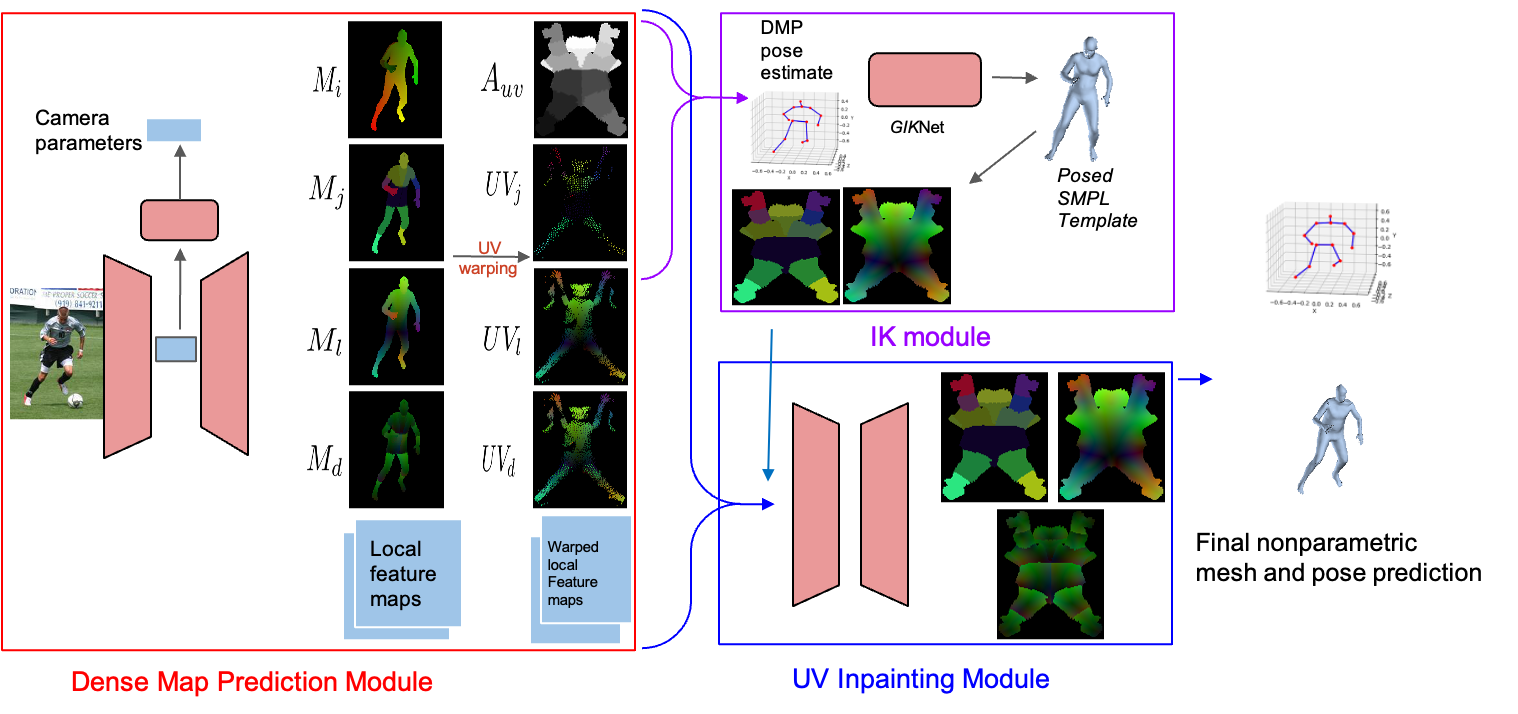}
\end{center}
   \caption{Our 3d body estimation framework consists of three part: Dense Map Prediction module (\textit{DMP}), Inverse Kinematics and SMPL module (\textit{IK}) and UV Inpainting Module (\textit{UVI}).
   }
\label{fig:flowchart}
\end{figure*}

\paragraph{Occlusion} \cite{howrobust}
presented a systematic study of various types of synthetic
occlusions in 3D human pose estimation from a single RGB
image. Since synthetic data can not fully depict the real occlusion, \cite{objectdetectiongrammarmodels} learns from real data and uses
grammar models with explicit occluding templates to reason about occluded people. To avoid specific design for
occlusion patterns, \cite{parseoccludepeople} presents a method for modeling occlusion that aims at explicitly learning the appearance
and statistics of occlusion patterns. They also synthesizes a large corpus of training data by compositing segmented objects at random locations over a base training image. \cite{occlusionawarevideo} utilizes a cylinder model and confidence maps to filter out the occluded joints and uses flow warped joint in the same video to approximate the missing joints. \cite{depthocclusion} integrates depth
information about occluded objects into 3D pose estimation. To provide full-geometry information to handle occlusion scenarios, \cite{gpa} and \cite{prox} provide 3d scene geometry as multi-layer depth maps or signed distance fields into the inference stage. \cite{partialobservation} proposes a simple but effeive self-training framework to adapt the model to highly occluded observations. To fully utilize the holistic human body model (e.g. SMPL \cite{smpl}), \cite{3DOH} represents the target SMPL human mesh as UV location map and converts the full-body human estimation as an image inpainting problem. However, these frameworks either rely on nonparametric estimation or pure model-based regression, how to leverage the best of both worlds seamlessly remain an unexplored problem.

\section{Method}
As shown in Fig~\ref{fig:flowchart}, our framework consists of three consecutive modules, including a dense map prediction module (\textit{DMP}), which extract dense semantic maps (e.g. 3d joint location, surface location and their displacements) and correspondence UV position, an inverse kinematics and SMPL module (\textit{IK}), which inpaint 3d joint location and estimate the smpl parameters, as well as a UV map inpainting module, which estimate the final joint location and mesh location in UV space.

\subsection{Dense Map Prediction Module}

\begin{figure}
\centering
\includegraphics[width=0.24\linewidth]{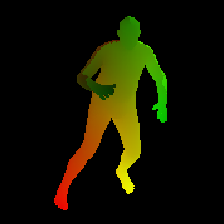}
\includegraphics[width=0.24\linewidth]{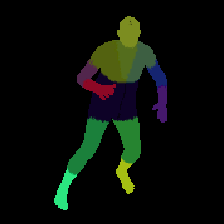}
\includegraphics[width=0.24\linewidth]{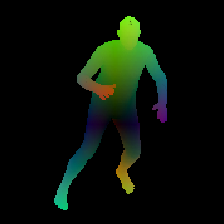}
\includegraphics[width=0.24\linewidth]{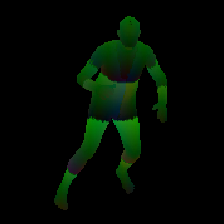}
\caption{{Semantic maps aligned with image space. From left to right: IUV image $M_{i}$, Dense jointmap $M_{j}$, dense location map $M_{l}$ and dense displacement map $M_{d}$. (Best viewed in Color)}}
\label{fig:singlesample}
\end{figure}

\begin{figure}
\centering
\includegraphics[width=0.24\linewidth]{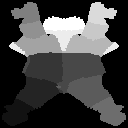}
\includegraphics[width=0.24\linewidth]{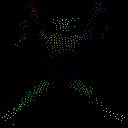}
\includegraphics[width=0.24\linewidth]{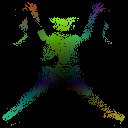}
\includegraphics[width=0.24\linewidth]{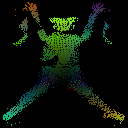}
\caption{{Warped Images in UV space based on IUV images $M_i$. From left to right: Part segmentation in UV space $A_{uv}$, UV space jointmap $UV_{j}$, UV space location map $UV_{l}$ and UV space displacement map $UV_{d}$. (Best viewed in Color) }}
\label{fig:singlesamplewarped}
\end{figure}

Our dense map prediction module is an encoder-decoder architecture and is used to extract the IUV images $M_i$, as well as dense semantic maps including dense joint map $M_j$, dense location map $M_l$ and dense displacement maps $M_d$. They are further illustrated in Fig~\ref{fig:singlesample}.  $M_i$ is generated from the continuous UV map from \cite{decomr}, it is continuous in both image space and UV space, thus, easier to learn compared with original UV map \cite{smpl}. It is used to convert the dense local features as well as these semantic maps to UV space. For location map $M_l$, it represents the position of each vertices from the SMPL human mesh surface in root-relative coordinates. To construct $M_l$ groundtruth, we first use the SMPL model, SMPL parameters and camera parameters to generate the vertices location in root-relative coordinate, and generate the full UV space location map $UV_l$ using barycentric interpolation (The mesh faces correspondence is defined by \cite{decomr}). After that we use the $M_i$ to fetch values from $UV_l$ to get the dense location map in image space.  For the generation of dense joint map $M_j$, we first rely on T-pose SMPL mesh and assign each vertex to the nearest joints (14 LSP joints setting), after that we use barycentric interpolation to get the UV space assignment, and further refine the assignment by make it symmetric in UV space (e.g. left hip and right hip has symmetric shape in UV space, as illustrated in Fig~\ref{fig:singlesamplewarped}). We term the part assignment in UV space  as $A_{uv}$ . After setting the assignment in UV space, we use the $M_i$ to query values from $UV_j$ to get the dense joint map in image space. $UV_j$ stores the root-relative joint location. We define displacement as the residual between vertex location and the assigned joint location, thus  $UV_d = UV_l - UV_j$ and $M_d = M_l - M_j$. As our human are left-right symmetric (e.g. left hand has symmetric shape with right hand and the size and the distance between joint and surface is almost the same.), the magnitude of left part and right part of $UV_d$ should be the same. 

These semantic maps are aligned with the human in the images. Thus we are able to train a encoder-decoder network to estimate directly from image space. Dense image space joint prediction shares the similar flavor with \cite{A2J,personlab}. 

The objective for the dense map prediction module is 

\begin{equation}
{\ell_{DMP}} = {\ell_{M_i}} + {\ell_{M_l}} + {\ell_{M_j}} + {\ell_{M_d}} 
  \label{eqn:DMP}
\end{equation}

$\ell_{M_i}$ is composed of two parts: a binary mask loss $\ell_{M_ib}$ of human body, which distinguishes pixels from those at the background, and the human pixels. The loss function of $\ell_{M_ib}$ is binary cross entropy loss.  our CNN further outputs the UV coordinates and uses L1 loss $\ell_{M_{iuv}}$ . 

\begin{equation}
{\ell_{M_i}} = {\ell_{M_{ib}}} +   {\ell_{M_{iuv}}}
  \label{eqn:mi}
\end{equation}

For $\ell_{M_l}$,  $\ell_{M_j}$ and  $\ell_{M_d}$, we use L1 loss to directly regress the real value. As these values are already in root-relative coordinate and in unit meters, thus their data range is $-1$ to $+1$, we do not further normalize them.  

Our dense map prediction module not only predicts these semantic maps, but also extracts both global feature to estimate camera parameter and local feature for the UV impainting module.

\begin{figure}
\centering
\includegraphics[width=0.27\linewidth]{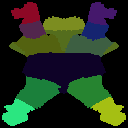}
\includegraphics[width=0.27\linewidth]{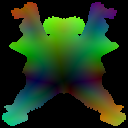}
\includegraphics[width=0.27\linewidth]{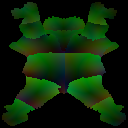}
\caption{{Full groundtruth in UV space. From left to right:  UV space jointmap $UV_{j}$, UV space location map $UV_{l}$ and UV space displacement map $UV_{d}$. (Best viewed in Color) }}
\label{fig:singlesamplewarpedfull}
\end{figure}

\subsection{Inverse Kinematics Module}
\paragraph{Estimate Joint Location from \textit{DMP}} After warping the semantic maps ($M_l,M_j,M_d$) from image space to uv space, we get the incomplete uv joint map $UV_j$. Based on the uv space joint assignment $A_{uv}$ (as shown in Fig~\ref{fig:singlesamplewarped}), we aggregate the dense prediction $UV_j$ for each joint and average them if they are not fully occluded. Thus we have a coarse prediction for each joint $J_{initial}$.

\begin{figure}
\centering
\includegraphics[width=0.9\linewidth]{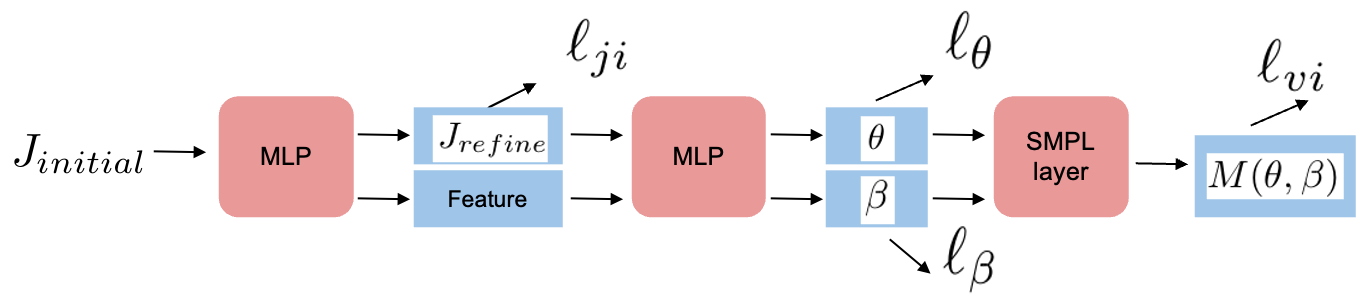}
\caption{{Structure of \textit{GIKNet}. (Best viewed in Color) }}
\label{fig:giknet}
\end{figure}

\paragraph{Joint Inpaint and Refine Module} Even though each human pixel contributes to joint prediction, there are still cases that some joints have no assigned vertex/pixel available from the image evidence. Thus we propose 
the joint inpainting module to inpaint these missing joints. This network is pretty flexible and can be MLP \cite{simple}, GCN \cite{semanticsgcn} or even modern transformers \cite{METRO}. For the ease of implementation we use simple multi-layer perceptron. Our joint inpainting net is inspired by \cite{simple}, which is simple, deep and a fully-connected network with six linear layer with 256 output features. It includes dropout after every fully connected layer, batch-normalization and residual connections. The model contains approximately 400k training parameters. The goal of this network is not only to inpaint the joints but also to refine the joints prediction that is not occluded. It takes  the $J_{initial}$ as input 
and the output of the network is the joint in root-relative coordinates $J_{refine}$. We use L1 loss $L_{ji}$ to train joint inpaint and refine module. The structure of the joint inpainting and refine module is shown in Fig~\ref{fig:giknet}. 

\paragraph{Inverse Kinematics Module} After getting the sparse 3d human keypoints. We want to repose the template SMPL meshes based on the predicted joints location. To solve this problem we leverage inverse kinematics (IK).  Typically, the IK task is tackled with iterative optimization methods \cite{ik1,ik2,ik3}, which requires a good initialization, more time and case-by-case optimization method. Here we propose a global inverse kinematics neural network  \textit{GIK-Net}. This network is constructed by the basic fully connected neural network module with residual connection, batch normalization and relu activation similar to \cite{simple}. In particular, \textit{GIK-Net} takes the refined keypoint coordinates  $J_{refine}$ in root-relative space and outputs joint rotations $\theta$ and $\beta$ which serve as the input for SMPL layer. As we also use the Mocap dataset (AMASS \cite{AMASS:ICCV:2019}, SPIN\cite{spin} and AIST++ \cite{aist++}), our \textit{GIK-Net} can implicitly learn the realistic distribution of human kinematics rotation and human shape. The use of the additional Mocap dataset serves the same purpose as the factorized adversarial prior \cite{HMR}, variational human pose prior \cite{SMPL-X:2019} and motion discriminator \cite{kocabas2019vibe}. We use L1 loss $L_{\theta}$  and $L_{\beta}$ to train \textit{GIK-Net}. The structure of \textit{GIK-Net} is shown in Fig \ref{fig:giknet}.

\paragraph{SMPL revisits and Reposing Module} SMPL \cite{smpl} represents the body pose and shape by pose $\theta\in R^{72}$ and shape $\beta\in R^{10}$ parameter. Here we use the gender-neural shape model following previous work \cite{pare,HMR,spin}. Given these parameters, the SMPL module is a differentiable function that outputs a posed 3D mesh $M(\theta,\beta) \in R^{6890 \times 3}$. The 3D joint locations $J_{3D} = WM \in R^{J \times 3}$, while J are computed with a pretrained linear regressor $W$.
After getting the $\theta$ and $\beta$ from the  \textit{GIK-Net} we send them to SMPL layer to get the body mesh prediction.  

We also augment  the joints input for \textit{GIK-Net} from Mocap dataset with guassian noise and random synthetic occlusion (30\%). The augmentation helps our \textit{GIK-Net} generalize to more realistic noisy input.
We use L1 loss  $L_{vi}$ to train the mesh prediction from SMPL module.

The objective for the inverse kinematics and smpl module is 

\begin{equation}
{\ell_{IK}} = {\ell_{\theta}} + {\ell_{\beta}} + {\ell_{ji}} + {\ell_{vi}} 
  \label{eqn:IK}
\end{equation}

\subsection{UV Inpainting Module}

The goal of UV inpainting module is to regress 3d joint and mesh location directly based on the feature / semantic output ($UV_l,UV_j,UV_d$) from \textit{DMP} and semantic output ($UV_l,UV_j,UV_d$) from \textit{IK}.

\paragraph{Inevitable Fitting Error introduced by SMPL model and Joint regressor} The advantage of directly regressing joint/mesh location over model-based method is that model-based method will introduce intrinsic fitting error. Specifically, we use the SMPL layer, groundtruth SMPL parameters (from Mosh),  and the joint-regressor \cite{spin} to obtain  fitted joint for the whole Human3.6M dataset. We get average fitting error as 24.1 mm (MPJPE) when compared with the Human3.6M joint from Mocap system. It means that even we predict perfect SMPL mesh we still have about 24.1 mm fitting error. Thus we argue directly train and estimate joint location from UV space is a better alternative solution. 

\paragraph{UV inpainting module} After getting the refined joint location $J_{refine}$ from \textit{IK} module, we   distribute the refined joint location in UV space based on UV space joint assignment map $A_{uv}$ and generate refine UV joint map $UV_{jrefine}$. We also have the reposed template mesh and the corresponding reposed UV location map $UV_{l}$ (through barycentric interpolation). 
Additionally, we have features $UV_f$, location map $UV_l$, joint map $UV_j$ and displacement $UV_d$ from \textit{DMP}. We combine the best of both worlds ( \textit{DMP} and  \textit{IK}) feature through aggregation and send it to our UV inpainting module. The UV inpainting module is a light UNet with skip connections. We can see the Fig \ref{fig:singlesamplewarpedfull} is the complete version of Fig \ref{fig:singlesamplewarped} and  serves as the groundtruth for the UV inpainting module.

For the training of the UV inpainting module, we have 
\begin{equation}
{\ell_{map}} = \| {\hat {UV}_{map}} - UV_{map} \|_{1}
  \label{eqn:map}
\end{equation}
Note the `map' represents  location map,  joint map and  displacement map in uv space. Addtionally, we have 3d joints and 2d joint loss based on the predicted camera parameter. Our camera parameters  consist of scale and offset parameter to map the xy in $J_{3d}$  to $J_{2d}$.
\begin{equation}
{\ell_{j3d}} = \| {\hat {J}_{3d}} - J_{3d} \|_{1}
  \label{eqn:j3d}
\end{equation}
\begin{equation}
{\ell_{j2d}} = \| {\hat {J}_{2d}} - J_{2d} \|_{1}
  \label{eqn:j2d}
\end{equation}
As we know, the distance between the human surface to the joints are left-right symmetric, thus we also apply symmetric loss on the magnitude of displacement.
\begin{equation}
{\ell_{dismag}} = \|  \|{\hat {UV}_{d}}\| - \|{\hat {UV}^{flip}_{d}}\| \|_{1}
  \label{eqn:displacementmagnitude}
\end{equation}

To align the predicted mesh surface with image aligned IUV images $M_i$, we also adopt consistent loss from \cite{decomr}. It is enabled by the camera parameter predicted by our model (scaling and offset parameter). 

The objective for the uv inpainting module is 

\begin{equation}
{\ell_{UVI}} = {\ell_{dismag}} + {\ell_{j2d}} + {\ell_{j3d}} + {\ell_{map}}  + {\ell_{con}}
  \label{eqn:UVI}
\end{equation}

Thus we have all the losses as 

\begin{equation}
{\ell_{all}} = {\ell_{DMP}} + {\ell_{IK}} + {\ell_{UVI}}
  \label{eqn:allloss}
\end{equation}

\paragraph{Inference} We do inference of 3d joint location from  $UV_j$ and based on the uv assignment $A_{uv}$ for each joint. We average all the prediction for the specific joints if this pixel prediction is valid. For human mesh prediction we use the barycentric interpolation from the UV space location map $UV_l$.

\subsection{Implementation Details}

The proposed framework is trained on the ResNet-50 \cite{resnet} backbone pre-trained on ImageNet. It takes a 224 $\times$ 224 image as input, and input resolution for  \textit{UVI}  is 64 $\times$ 64 and the output resolution is 128 $\times$ 128. 
We train three modules separately. We first train our \textit{DMP}, and based on the output of  \textit{DMP} and Mocap data we train our \textit{IK}; We finally fix and concat \textit{DMP} and \textit{IK}, and train \textit{UVI} module. We apply synthetic occlusion \cite{SyntheticOcclusion} when train \textit{DMP}. The training data is augmented with randomly scaling, rotation, flipping and RGB channel noise. We use the Adam optimizer. The training data for each module is illustrated in Table \ref{table:traingset}.

\section{Experiments}

\subsection{Dataset and Evaluation Metric}
 
\begin{table}
\begin{center}
\small
\begin{tabular}{l|c}
\hline
Stages  & Training Datasets  \\
\hline
\textit{DMP}  & H36M, MPI-INF-3DHP, MPII, COCO, LSP\\ 
\textit{IK}   & H36M, MPI-INF-3DHP, AMASS, AIST++     \\ 
\textit{UVI}   & H36M, 3DOH \\ 
\hline
\end{tabular}
\end{center}
\caption{Training datasets for each module.}
\label{table:traingset}
\end{table}

 \paragraph{Human3.6M} \cite{h36m_pami}  is commonly used as the benchmark
dataset for 3D human pose estimation, consisting of 3.6 millions of video frames captured in the controlled environment. It has 11 subjects, 15 kinds of action sequences and 1.5 million training images with accurate 3D
annotations. 
Similar to \cite{HMR}, we use MoSH to process the marker data in the original dataset, and obtain the ground truth SMPL parameters to generate the groundtruth for $UV_l$. For a fair comparison, we use 300K data in S1, S5, S6, S7, S8 for network training, and test in S9, S11. 
\\
\textbf{3DOH} \cite{3DOH} utilize multi-view SMPLify-X \cite{SMPL-X:2019} to get the 3d ground truth. The dataset is designed to have object occlusion for subjects. It contains 50,310 training images and 1,290
test images. It provides 2D, 3D annotations and SMPL parameters to generate meshes. We use the test set for evaluation purposes and the training set to train the \textit{UVI} module. 
\\
\textbf{LSP} \cite{lsp} dataset is a 2D human pose estimation
benchmark. In our work, we use the \cite{up3d} SMPL parameter to render the $M_{i}$ to train \textit{DMP} module.
\\
\textbf{MPI-INF-3DHP} \cite{mono_3dhp2017} is a dataset captured with a multi-view
setup mostly in indoor environments. No markers are used for the capture, so 3D pose data tend to be less accurate compared to other datasets. We use the provided training
set (subjects S1 to S8) for training. We use the it to train \textit{DMP} and \textit{IK} module.
\\
\textbf{Mocap dataset} We use \cite{AMASS:ICCV:2019} AMASS, AIST++ \cite{aist++} and SPIN \cite{spin} dataset to train our occlusion aware \textit{GIKNet}. 
\\
\textbf{Evaluation} We evaluate our method on H36M \cite{h36m_pami} dataset and  3DOH \cite{3DOH} datasets. We report Procrustes-aligned mean per joint position error (MPJPE-PA) and mean per joint position error (MPJPE) in mm. For 3DOH we also report mean per vertex error (MPVE) in mm.

\begin{figure}
\centering
\includegraphics[width=0.35\linewidth]{00001_image_71_102_uvspace_seg_addone_uvassign.png}
\includegraphics[width=0.35\linewidth]{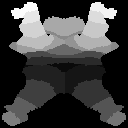}
\caption{{Different part segmentation choice in UV space. (Best viewed in Color) }}
\label{fig:uvpartseg14and24}
\end{figure}

\begin{table}
\begin{center}
\small
\begin{tabular}{l |c|c }
\hline
        &    \multicolumn{2}{c}{H36M} \\
\hline
Method  &  MPJPE & MPJPE-PA \\
\hline
HMR \cite{HMR}    & 88.0   & 56.8  \\
DaNet \cite{pamigcndensepose} & 61.5   & 48.6  \\
HoloPose \cite{holopose}  & 60.3   & 46.5  \\
SPIN \cite{spin}    & 62.5   & 41.1  \\
I2L \cite{I2L}     &  \underline{55.7}  &  41.1 \\
DetNet \cite{zhou2021monocular}  & 64.8 & 50.3 \\
PHMR \cite{HybrIK}     &  -  &  41.2 \\
DecoMR \cite{decomr}      &  60.5  &  \underline{39.3} \\
PyMaf \cite{pymaf}      &  57.7  &  40.5 \\
\hline
Ours \textit{DMP}-14    & 69.7 & 51.7  \\
Ours \textit{IK}-14       &67.3 & 50.6\\
Ours \textit{UVI}-14       & \textbf{54.7}  & \textbf{38.4} \\
\hline
\end{tabular}
\end{center}
\caption{Reconstruction errors on Human3.6M dataset.}
\label{table:h36m}
\end{table}



\begin{table}
\begin{center}
\small
\begin{tabular}{l|c|c |c}
\hline
      &  \multicolumn{3}{c}{3DOH}  \\
\hline
Method  & MPJPE & MPJPE-PA & MPVE\\
\hline
SMPLify-X  &  -  & 156.4   &  177.3\\
OOH \cite{cvprooh}  &  -  & 58.5   &  \textbf{63.3}\\
SPIN \cite{spin}   &   104.3 & 68.3 & 113.4\\
PyMAF\cite{pymaf}   &   96.2 & - & 107.3\\
HMR-EFT$^{\star}$ \cite{EFT} &   75.2 & 53.1 & -\\
PARE$^{\star}$ \cite{pare}  &  \underline{63.3} & \textbf{44.3}  & -\\
\hline
Ours \textit{DMP}-14    & 128.4 & 109.8  & - \\
Ours \textit{IK}-14    & 112.9  & 80.8 & 133.5\\
Ours \textit{UVI}-14    &  \textbf{58.3} &  \underline{44.6} &  \underline{72.3}   \\
\hline
\end{tabular}
\end{center}
\caption{Comparison with SOTA performance on 3DOH dataset. $\star$ denotes the model trained on better ground truth data from EFT \cite{EFT}. }
\label{table:3doh}
\end{table}


\begin{figure*}[t]
\begin{center}
   \includegraphics[width=0.92\linewidth]{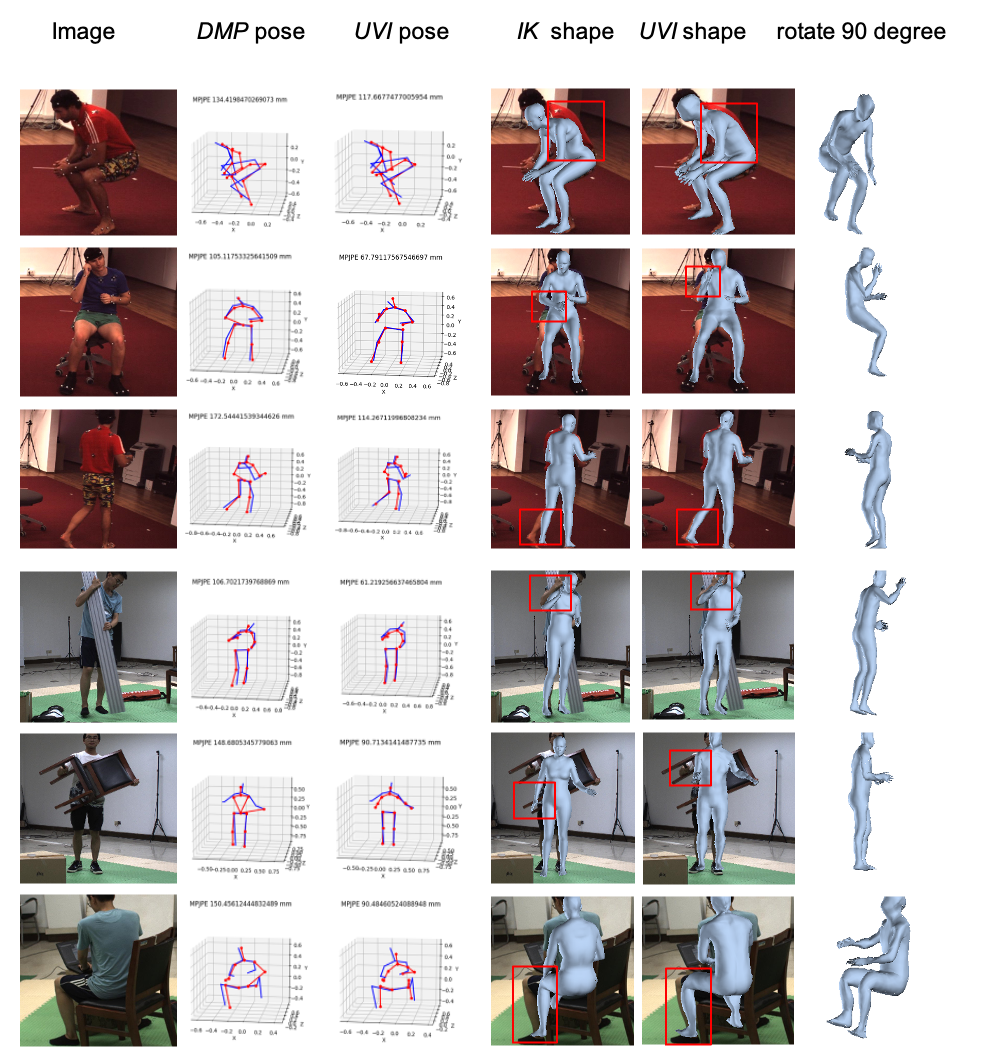}
\end{center}
   \caption{Pose and shape prediction from \textit{DMP} module, \textit{IK} module and \textit{UVI} module. (Best viewed in Color) }
\label{fig:qualitative}
\end{figure*}




\subsection{Quantitative Results}

\paragraph{Comparison with SOTA performance} We can see our final stage (\textit{UVI}-14) in Table~\ref{table:h36m} achieve state-of-the-art performance on common H36M benchmark. Our SOTA performance demonstrates the usefulness of proposed combination of model-based and nonparametric approaches. In Table~\ref{table:3doh}, as our methods focus on both pose and mesh while \cite{cvprooh} focus more on meshes, they achieve SOTA performance on 3DOH dataset; PARE \cite{pare} uses the  EFT dataset \cite{EFT} with improved groundtruth  thus outperforms us on MPJPE-PA metric.  

\paragraph{14 joints vs 24 joints setting} Another way to get 24 joints prediction from \textit{DMP} is to have a 24 joints segmentation $A_{uv}$ in UV space following SMPL setting. As shown in Fig~\ref{fig:uvpartseg14and24} we define 14 joints setting and 24 joints setting. We run \textit{DMP}-24 and  \textit{DMP}-14 and evaluate on the predicted $J_{initial}$. We observe the error of \textit{DMP}-24 is much higher than  \textit{DMP}-14 as in Table \ref{table:3dohablstions}. The main reason is that over-segment of body parts may distribute less visible pixels to certain parts (feet, hand) and will lead to  higher error.

\paragraph{Occlusion vs Non-occlusion} When computing the MPJPE for $J_{initial}$, the results for visible parts (part with any pixel belong to them visible) and invisible parts  differ a lot. We compare the \textit{DMP-14} and \textit{DMP-14-Nonoccluded} in Table \ref{table:3dohablstions}. We find visible parts with 87.3 mm MPJPE while the MPJPE counting invisible parts yield 128.4 mm. It tells us if the joints are visible, our \textit{DMP} can predict relative good initial results. Thus, synthetic occlusion helps for  our \textit{DMP} module.
When we remove the data augmentation techniques like synthetic occlusion \cite{SyntheticOcclusion}, \textit{DMP-14} increase to 135.4 mm.

\begin{table}
\begin{center}
\small
\begin{tabular}{l |c|c|c }
\hline
        &  \multicolumn{3}{c}{3DOH}  \\
\hline
Method  &  MPVE & MPJPE & PMPJPE  \\
\hline
\textit{DMP}-24    & -  & 246.4   & 208.5  \\
\textit{DMP}-14  w/o synthetic occlusion  & -  &  135.4 & 115.7 \\
\textit{DMP}-14-Nonoccluded    & -  &  87.3 & 64.7 \\
\textit{DMP}-14    & -  &  128.4 & 109.8 \\
\hline

\textit{IK}-14 w/o gaussian noise    & 138.2  &  115.7  & 82.8 \\
\textit{IK}-14 w/o random zero    & 139.5  &  116.8  & 83.2 \\
\textit{IK}-14     & 133.5  &  112.9  & 80.8 \\
\hline
\textit{UVI}-14  w/o \textit{IK}-14   &   82.9 &  69.4 & 58.1  \\
\textit{UVI}-14  w/o \textit{DMP}-14   &   80.1  &   67.8 &  55.1  \\
\textit{UVI}-14  w/o $\ell_{dismag}$   & 75.5   &  63.8  &  47.3 \\
\textit{UVI}-14     &  72.3  & 58.3   & 44.6  \\
\hline
\end{tabular}
\end{center}
\caption{Ablation study about reconstruction errors on 3DOH test set. 14 and 24 denotes the number of joints setting for training and evaluations. Nonoccluded denotes when we calculate error we are not counting the part without any visible image evidence.}
\label{table:3dohablstions}
\end{table}

\paragraph{\textit{GIK-Net} data augmentations} We also try to remove the gaussian noise or random mask out joints data augmentation techniques for MOCAP data, which serve as input for the \textit{GIK-Net}, to see how is the MPJPE varying. 
As shown in Table \ref{table:3dohablstions}, \textit{IK-14 w/o gaussian noise} and \textit{IK-14-w/o random zero} yield larger error (2.8 mm and 3.9 mm ) compared with \textit{IK-14}. It demonstrate these data augmentation makes the \textit{GIK-Net} more robust to noise and helps generalize to real data input.  

\paragraph{\textit{UVI} ablations} As the magnitude of our $UV_{d}$ should be symmetric, we introduce the magnitude error for $UV_{d}$ and its flip version.  We run a model without this $\ell_{dismag}$ and  observe there is 4.5 mm error increase in MPJPE metric. This is shown in Table \ref{table:3dohablstions}. 

\paragraph{Each stage performance} \textit{DMP} module is a nonparametric method, while \textit{IK} module is a model-based method relying on the output of \textit{DMP} and then correct it. \textit{UVI} module relies on both nonparametric output and model-based output, and predict the final body joint and mesh. Based on Table \ref{table:3dohablstions}, \textit{DMP-14} estimate from raw images and gives inferior performance.  \textit{IK}-14 corrects the output from \textit{DMP-14} and reduce the error by 15.5 mm. \textit{UVI}-14 relies on both \textit{IK-14} and \textit{DMP-14} and further reduce MPJPE to 58.3 mm. However, if any of the previous stage output is missing, MPJPE increase by 11.1 mm (w/o \textit{IK-14}) or 8.5 mm (w/o \textit{DMP-14}).






\subsection{Qualitative Results} 
We present qualitative results in Fig~\ref{fig:qualitative} including the joints prediction from \textit{DMP}, \textit{IK}, \textit{UVI} modules and mesh prediction from \textit{IK}, \textit{UVI} modules. 

\begin{figure}
\centering
\includegraphics[width=0.96\linewidth]{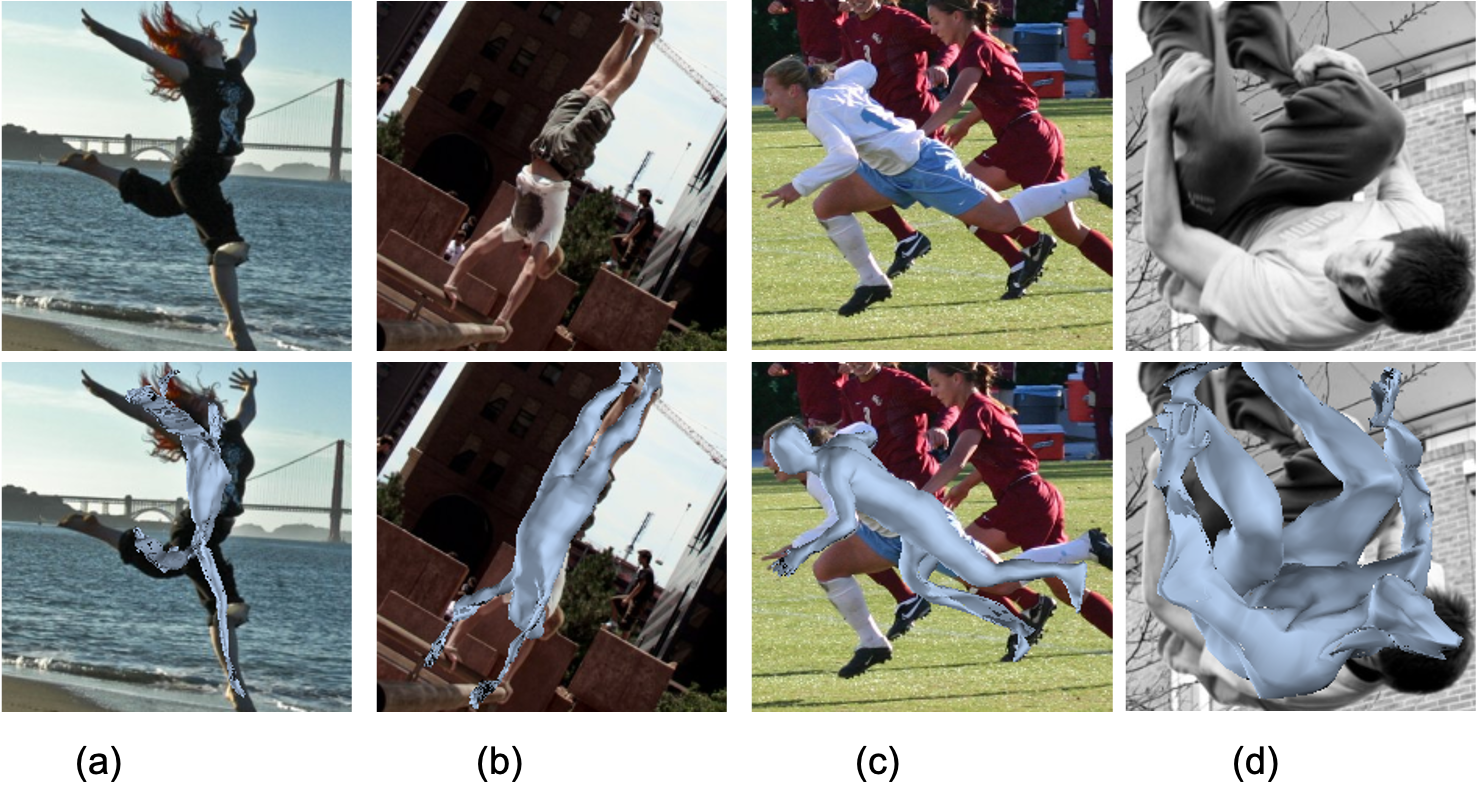}
\caption{Failure cases. (Best viewed in color) }
\label{fig:failurecases}
\end{figure}

\paragraph{Limitations} We also show failure cases in Fig \ref{fig:failurecases}. Typical failure cases can be attributed to challenging poses (a,b,d), and crowded scenarios (c). 

\section{Conclusion}
We propose a framework that combine the best of both worlds (nonparametric and SMPL model-based method). It predicts the initial 3d body pose from \textit{DMP} module, refine the predicted pose and repose the template SMPL meshes using \textit{IK} module. Based on the nonparametric prediction from \textit{DMP} module and model-based prediction from \textit{IK} module, the \textit{UVI} module inpaint and refine the prediction. To alleviate the intrinsic error introduced by joint regressor (fitting), we regress joint ($UV_j$) and mesh ($UV_l$) separately in different maps in UV space. We also introduce the magnitude loss $\ell_{dismag}$ to enforce the symmetric property of human ($UV_d$). Our framework achieves state-of-the-art performance among 3D
mesh-based methods on several public benchmarks. Future work can focus on extending the framework to the reconstruction of full body surfaces including hands and faces. 

{\small
\bibliographystyle{ieee_fullname}
\bibliography{main}

\begin{thebibliography}{10}\itemsep=-1pt

\bibitem{ik1}
A Balestrino, Giuseppe~De Maria, and L Sciavicco.
\newblock Robust control of robotic manipulators.
\newblock In {\em IFAC Proceedings Volumes}, 1984.

\bibitem{occlusionawarevideo}
Yu Cheng, Bo Yang, Bo Wang, Wending Yan, and Robby~T. Tan.
\newblock Occlusion-aware networks for 3d human pose estimation in video.
\newblock In {\em ICCV}, 2019.

\bibitem{parseoccludepeople}
Golnaz Ghiasi, Yi Yang, Deva Ramanan, and Charless~C. Fowlkes.
\newblock Parsing occluded people.
\newblock In {\em CVPR}, 2014.

\bibitem{ik2}
Michael Girard and Anthony~A Maciejewski.
\newblock Computational modeling for the computer animation of legged figures.
\newblock In {\em SIGGRAPH}, 1985.

\bibitem{objectdetectiongrammarmodels}
Ross~B. Girshick, Pedro~F. Felzenszwalb, and David McAllester.
\newblock Object detection with grammar models.
\newblock In {\em Neurips}, 2020.

\bibitem{holopose}
Riza~Alp Guler and Iasonas Kokkinos.
\newblock Holopose: Holistic 3d human reconstruction in-the-wild.
\newblock In {\em CVPR}, 2019.

\bibitem{prox}
Mohamed Hassan, Vasileios Choutas, Dimitrios Tzionas, and Michael~J. Black.
\newblock Resolving 3d human pose ambiguities with 3d scene constraints.
\newblock In {\em ICCV}, 2019.

\bibitem{resnet}
Kaiming He, Xiangyu Zhang, Shaoqing Ren, and Jian Sun.
\newblock Deep residual learning for image recognition.
\newblock In {\em CVPR}, 2016.

\bibitem{h36m_pami}
Catalin Ionescu, Dragos Papava, Vlad Olaru, and Cristian Sminchisescu.
\newblock Human3.6m: Large scale datasets and predictive methods for 3d human
  sensing in natural environments.
\newblock In {\em PAMI}, 2014.

\bibitem{lsp}
Sam Johnson and Mark Everingham.
\newblock Clustered pose and nonlinear appearance models for human pose
  estimation.
\newblock In {\em BMVC}, 2010.

\bibitem{EFT}
Hanbyul Joo, Natalia Neverova, and Andrea Vedaldi.
\newblock Exemplar fine-tuning for 3d human pose fitting towards in-thewild 3d
  human pose estimation.
\newblock In {\em Arxiv}, 2020.

\bibitem{HMR}
Angjoo Kanazawa, Michael~J. Black, David~W. Jacobs, and Jitendra Malik.
\newblock End-to-end recovery of human shape and pose.
\newblock In {\em CVPR}, 2018.

\bibitem{kocabas2019vibe}
Muhammed Kocabas, Nikos Athanasiou, and Michael~J. Black.
\newblock Vibe: Video inference for human body pose and shape estimation.
\newblock In {\em CVPR}, 2020.

\bibitem{pare}
Muhammed Kocabas, Chun-Hao~P. Huang, Otmar Hilliges, and Michael~J. Black.
\newblock Pare: Part attention regressor for 3d human body estimation.
\newblock In {\em ICCV}, 2021.

\bibitem{spin}
Nikos Kolotouros, Georgios Pavlakos, Michael~J Black, and Kostas Daniilidis.
\newblock Learning to reconstruct 3d human pose and shape via model-fitting in
  the loop.
\newblock In {\em ICCV}, 2019.

\bibitem{up3d}
Christoph Lassner, Javier Romero, Martin Kiefel, Federica Bogo, Michael~J.
  Black, and Peter~V. Gehler.
\newblock Unite the people – closing the loop between 3d and 2d human
  representations.
\newblock In {\em CVPR}, 2017.

\bibitem{HybrIK}
Jiefeng Li, Chao Xu, Zhicun Chen, Siyuan Bian, Lixin Yang, and Cewu Lu.
\newblock Hybrik: A hybrid analytical-neural inverse kinematics solution for 3d
  human pose and shape estimation.
\newblock In {\em CVPR}, 2021.

\bibitem{aist++}
Ruilong Li, Shan Yang, David~A. Ross, and Angjoo Kanazawa.
\newblock Ai choreographer: Music conditioned 3d dance generation with aist++.
\newblock In {\em ICCV}, 2021.

\bibitem{Meshtransformer}
Kevin Lin, Lijuan Wang, and Zicheng Liu.
\newblock End-to-end human pose and mesh reconstruction with transformers.
\newblock In {\em CVPR}, 2021.

\bibitem{METRO}
Kevin Lin, Lijuan Wang, and Zicheng Liu.
\newblock End-to-end human pose and mesh reconstruction with transformers.
\newblock In {\em CVPR}, 2021.

\bibitem{meshgraphormer}
Kevin Lin, Lijuan Wang, and Zicheng Liu.
\newblock Mesh graphormer.
\newblock In {\em ICCV}, 2021.

\bibitem{lin2014microsoft}
Tsung-Yi Lin, Michael Maire, Serge Belongie, James Hays, Pietro Perona, Deva
  Ramanan, Piotr Doll{\'a}r, and C~Lawrence Zitnick.
\newblock Microsoft coco: Common objects in context.
\newblock In {\em ECCV}, 2014.

\bibitem{smpl}
Matthew Loper, Naureen Mahmood, Javier Romero, Gerard Pons-Moll, and Michael~J.
  Black.
\newblock Smpl: A skinned multi-person linear model.
\newblock In {\em siggraph}, 2015.

\bibitem{pchmr}
Tianyu Luan, Yali Wang, Junhao Zhang, Zhe Wang, Zhipeng Zhou, and Yu Qiao.
\newblock Pc-hmr: Pose calibration for 3d human mesh recovery from 2d
  images/videos.
\newblock In {\em AAAI}, 2021.

\bibitem{haoyubmvc2021}
Haoyu Ma, Liangjian Chen, Deying Kong, Zhe Wang, Xingwei Liu, Hao Tang, Xiangyi
  Yan, Yusheng Xie, Shih-Yao Lin, and Xiaohui Xie.
\newblock Transfusion: Cross-view fusion with transformer for 3d human pose
  estimation.
\newblock In {\em BMVC}, 2021.

\bibitem{AMASS:ICCV:2019}
Naureen Mahmood, Nima Ghorbani, Nikolaus~F. Troje, Gerard Pons-Moll, and
  Michael~J. Black.
\newblock {AMASS}: Archive of motion capture as surface shapes.
\newblock In {\em ICCV}, 2019.

\bibitem{simple}
Julieta Martinez, Rayat Hossain, Javier Romero, and James~J. Little.
\newblock A simple yet effective baseline for 3d human pose estimation.
\newblock In {\em ICCV}, 2017.

\bibitem{mono_3dhp2017}
Dushyant Mehta, Helge Rhodin, Dan Casas, Pascal Fua, Oleksandr Sotnychenko,
  Weipeng Xu, and Christian Theobalt.
\newblock Monocular 3d human pose estimation in the wild using improved cnn
  supervision.
\newblock In {\em 3DV}, 2017.

\bibitem{rootnet}
Gyeongsik Moon, Juyong Chang, and Kyoung~Mu Lee.
\newblock Camera distance-aware top-down approach for 3d multi-person pose
  estimation from a single rgb image.
\newblock In {\em ICCV}, 2019.

\bibitem{I2L}
Gyeongsik Moon and Kyoung~Mu Lee.
\newblock I2l-meshnet: Image-to-lixel prediction network for accurate 3d human
  pose and mesh estimation from a single rgb image.
\newblock In {\em ECCV}, 2020.

\bibitem{personlab}
George Papandreou, Tyler Zhu, Liang-Chieh Chen, Spyros Gidaris, Jonathan
  Tompson, and Kevin Murphy.
\newblock Personlab: Person pose estimation and instance segmentation with a
  bottom-up, part-based, geometric embedding model.
\newblock In {\em ECCV}, 2018.

\bibitem{SMPL-X:2019}
Georgios Pavlakos, Vasileios Choutas, Nima Ghorbani, Timo Bolkart, Ahmed A.~A.
  Osman, Dimitrios Tzionas, and Michael~J. Black.
\newblock Expressive body capture: 3d hands, face, and body from a single
  image.
\newblock In {\em CVPR}, 2019.

\bibitem{volumetric}
Georgios Pavlakos, Xiaowei Zhou, Konstantinos~G Derpanis, and Kostas
  Daniilidis.
\newblock Coarse-to-fine volumetric prediction for single-image 3{D} human
  pose.
\newblock In {\em CVPR}, 2017.

\bibitem{depthocclusion}
Umer Raf, Juergen Gall, and Bastian Leibe.
\newblock A semantic occlusion model for human pose estimation from a single
  depth image.
\newblock In {\em CVPRW}, 2015.

\bibitem{partialobservation}
Chris Rockwell and David Fouhey.
\newblock Full-body awareness from partial observations.
\newblock In {\em ECCV}, 2020.

\bibitem{LCRnet++}
Gregory Rogez, Philippe Weinzaepfel, and Cordelia Schmid.
\newblock Lcr-net++: Multi-person 2d and 3d pose detection in natural images.
\newblock In {\em PAMI}, 2019.

\bibitem{integral}
Xiao Sun, Bin Xiao, Fangyin Wei, Shuang Liang, and Yichen Wei.
\newblock Integral human pose regression.
\newblock In {\em ECCV}, 2018.

\bibitem{howrobust}
István Sárándi, Timm Linder, Kai~O. Arras, and Bastian Leibe.
\newblock How robust is 3d human pose estimation to occlusion?
\newblock In {\em Arxiv}, 2018.

\bibitem{SyntheticOcclusion}
István Sárándi, Timm Linder, Kai~O. Arras, and Bastian Leibe.
\newblock Synthetic occlusion augmentation with volumetric heatmaps for the
  2018 eccv posetrack challenge on 3d human pose estimation.
\newblock In {\em Arxiv}, 2018.

\bibitem{motionretarget}
Ruben Villegas, Jimei Yang, Duygu Ceylan, and Honglak Lee.
\newblock Neural kinematic networks for unsupervised motion retargetting.
\newblock In {\em CVPR}, 2018.

\bibitem{gpa}
Zhe Wang, Liyan Chen, Shaurya Rathore, Daeyun Shin, and Charless Fowlkes.
\newblock Geometric pose affordance: 3d human pose with scene constraints.
\newblock In {\em arxiv}, 2019.

\bibitem{cross}
Zhe Wang, Daeyun Shin, and Charless Fowlkes.
\newblock Predicting camera viewpoint improves cross-dataset generalization for
  3d human pose estimation.
\newblock In {\em ECCV 3DPW workshop}, 2020.

\bibitem{ik3}
William~A Wolovich and H Elliott.
\newblock A computational technique for inverse kinematics.
\newblock In {\em CDC}, 1984.

\bibitem{xiao2018simple}
Bin Xiao, Haiping Wu, and Yichen Wei.
\newblock Simple baselines for human pose estimation and tracking.
\newblock In {\em ECCV}, 2018.

\bibitem{A2J}
Fu Xiong, Boshen Zhang, Yang Xiao, Zhiguo Cao, Taidong Yu, Joey~Tianyi Zhou,
  and Junsong Yuan.
\newblock A2j: Anchor-to-joint regression network for 3d articulated pose
  estimation from a single depth image.
\newblock In {\em ICCV}, 2019.

\bibitem{decomr}
Wang Zeng, Wanli Ouyang, Ping Luo, Wentao Liu, and Xiaogang Wang.
\newblock 3d human mesh regression with dense correspondence.
\newblock In {\em CVPR}, 2020.

\bibitem{cvprooh}
Tianshu Zhan, Buzhen Huang, and Yangang Wangu.
\newblock Object-occluded human shape and pose estimation from a single color
  image.
\newblock In {\em CVPR}, 2020.

\bibitem{pamigcndensepose}
Hongwen Zhang, Jie Cao, Guo Lu, Wanli Ouyang, and Zhenan Sun.
\newblock Learning 3d human shape and pose from dense body parts.
\newblock {\em PAMI}, 2020.

\bibitem{pymaf}
Hongwen Zhang, Yating Tian, Xinchi Zhou, Wanli Ouyang, Yebin Liu, Limin Wang,
  and Zhenan Sun.
\newblock Pymaf: 3d human pose and shape regression with pyramidal mesh
  alignment feedback loop.
\newblock In {\em ICCV}, 2021.

\bibitem{dynamicpose}
Junhao Zhang, Yali Wang, Zhipeng Zhou, Tianyu Luan, Zhe Wang, and Yu Qiao.
\newblock Learning dynamical human-joint affinity for 3d pose estimation in
  videos.
\newblock In {\em TIP}, 2021.

\bibitem{spatialarrangements}
Jason~Y. Zhang, Sam Pepose, Hanbyul Joo, Deva Ramanan, Jitendra Malik, and
  Angjoo Kanazawa.
\newblock Perceiving 3d human-object spatial arrangements from a single image
  in the wild.
\newblock In {\em ECCV}, 2020.

\bibitem{3DOH}
Tianshu Zhang, Buzhen Huang, and Yangang Wang.
\newblock Object-occluded human shape and pose estimation from a single color
  image.
\newblock In {\em CVPR}, 2020.

\bibitem{semanticsgcn}
Long Zhao, Xi Peng, Yu Tian, Mubbasir Kapadia, and Dimitris~N. Metaxas.
\newblock Semantic graph convolutional networks for 3d human pose regression.
\newblock In {\em CVPR}, 2019.

\bibitem{zhou2021monocular}
Yuxiao Zhou, Marc Habermann, Ikhsanul Habibie, Ayush Tewari, Christian
  Theobalt, and Feng Xu.
\newblock Monocular real-time full body capture with inter-part correlations.
\newblock In {\em CVPR}, 2021.

\bibitem{zhou2020monocular}
Yuxiao Zhou, Marc Habermann, Weipeng Xu, Ikhsanul Habibie, Christian Theobalt,
  and Feng Xu.
\newblock Monocular real-time hand shape and motion capture using multi-modal
  data.
\newblock In {\em CVPR}, 2020.

\end{thebibliography}
}

\newpage

\appendix
\section*{Appendix}
\section{Qualitative Results}

\begin{figure*}
\centering
\includegraphics[width=0.96\linewidth]{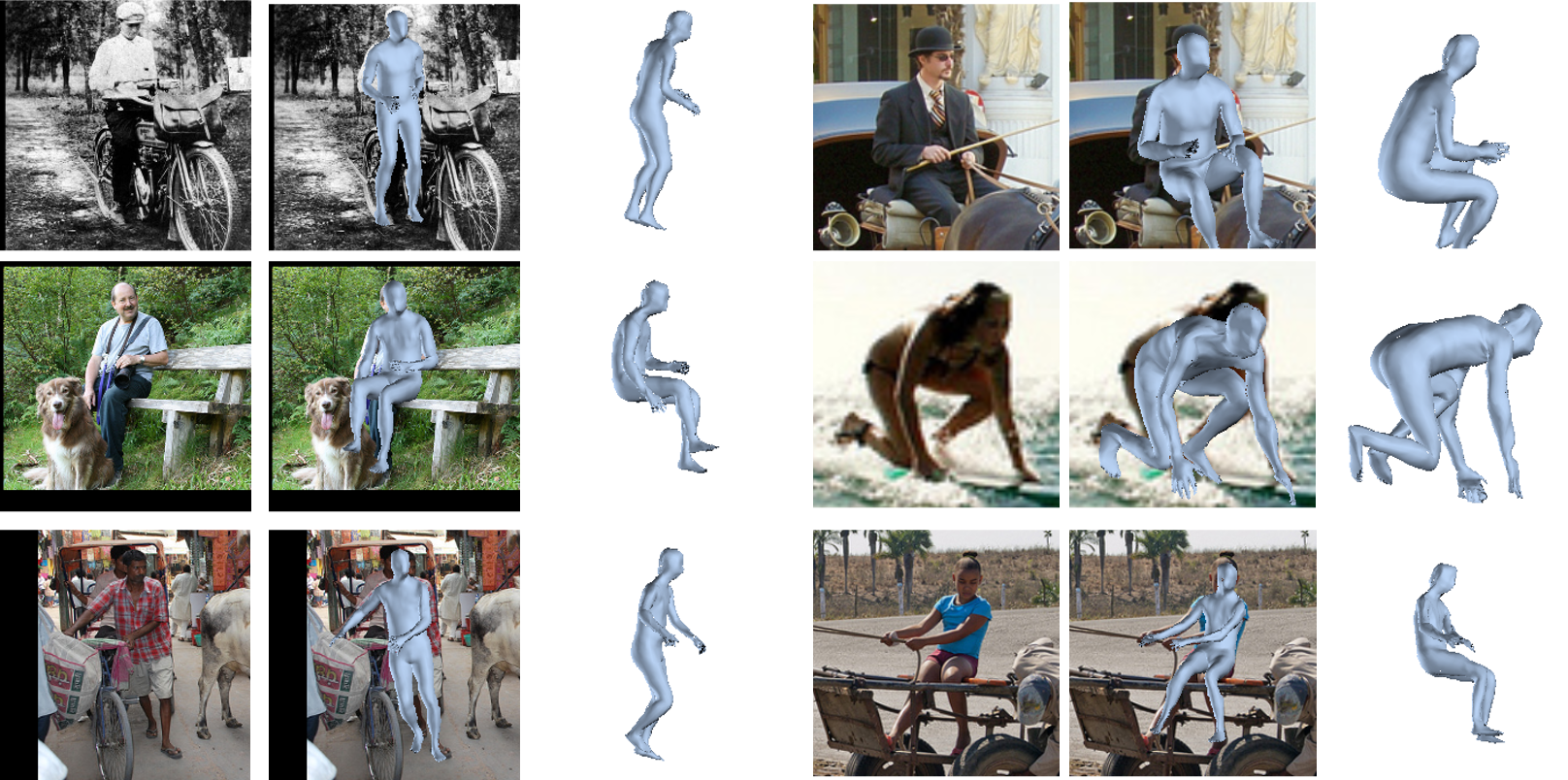}
\caption{{More qualitative results on COCO dataset. (Best viewed in Color) }}
\label{fig:morecoco}
\end{figure*}

\begin{figure*}
\centering
\includegraphics[width=0.96\linewidth]{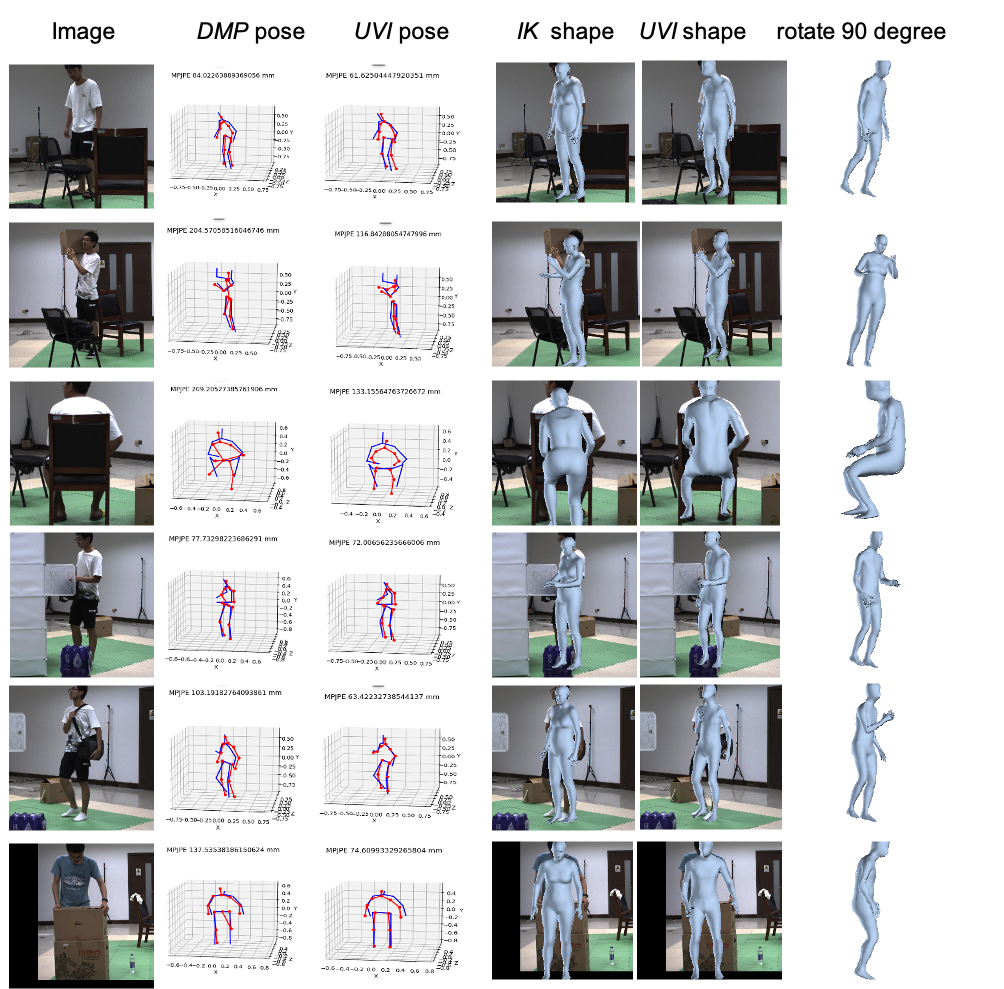}
\caption{{More qualitative results on 3DOH dataset. (Best viewed in Color) }}
\label{fig:more3doh}
\end{figure*}

We show more qualitative results on COCO \cite{lin2014microsoft} in Fig \ref{fig:morecoco}, and 3DOH \cite{3DOH} in Fig \ref{fig:more3doh}.

\section{Part Segmentation in UV Space}

We first use the reference T-pose mesh and the LSP joint regressor provided by \cite{spin,decomr} to get the T-pose 14 joint location. Then we calculate the joint-vertex euclidean distance and assign vertex to joint based on the smallest distance. After that, we use the barycentric interpolation (mapping between vertex triangle and UV space triangles) to get the UV space assignment probability ($128 \times 128 \times 14$). Following these operations, we use argmax to get the final assignment for each UV grid to the joint location. 

\section{Implementation Details}

For AMASS \cite{AMASS:ICCV:2019} data, we only get SMPL-H \cite{SMPL-X:2019} fitting instead of SMPL fitting data, however, SMPL-H does not included hands rotations as in SMPL. We sample random rotations from SPIN \cite{spin} fitting or the predictions from our \textit{DMP} stages for its training data. For AIST++ \cite{aist++}, it does not included $\beta$ parameters, we sample $\beta$ from SPIN \cite{spin} fitting or the predictions from our \textit{DMP} stages for its training data. We use the original rotation representation from SMPL \cite{smpl} (axis-angle representation) for the fast training purpose. 

\end{document}